%% file: acl_public.tex
\documentclass[11pt]{article}
\ifdefined\pdfminorversion\pdfminorversion=7\fi

\usepackage[final]{acl}

\usepackage{times}
\usepackage{latexsym}

\usepackage[T1]{fontenc}

\usepackage[utf8]{inputenc}
\usepackage{booktabs}
\usepackage{makecell}
\usepackage{xspace}
\usepackage{microtype}

\usepackage{inconsolata}
\usepackage{xspace}
\usepackage{amsmath}

\usepackage{graphicx}
\usepackage{booktabs}

\usepackage{array}
\usepackage{tabularx}
\usepackage{fontawesome5}
\newcommand{\sysname}{\textsc{SlideLab}\xspace}

\usepackage{subcaption}
\usepackage{xurl}

\title{\sysname: Audience-Centered Scientific Slide Generation and Evaluation}

\author{
\textbf{Vidushee Vats}\thanks{Equal contribution.}\thanks{Work done during internship at INSAIT.},
\textbf{Karun Sharma}\footnotemark[1]\footnotemark[2],
\textbf{Yuxia Wang} \\
INSAIT, Sofia University ``St. Kliment Ohridski'' \\
\vspace{1mm}
\texttt{vatsvidushee@gmail.com, mailto.karun.py@gmail.com} \\
\texttt{yuxia.wang@insait.ai}
}

\usepackage{booktabs}   \usepackage{xcolor}     \usepackage{amssymb}    \newcommand{\yes}{\textcolor{green!50!black}{\checkmark}}
\newcommand{\no}{\textcolor{red!70!black}{$\times$}}
\newcommand{\none}{\textcolor{gray!55}{$\circ$}}

\begin{document}
\maketitle
\begin{abstract}
Scientific presentations are more than summaries of research papers. They need to present the work in a coherent sequence, explain the main ideas clearly, and help the audience follow the presentation. We present \sysname, a training-free multi-agent framework for generating scientific presentations from research papers. \sysname first plans the presentation narrative, then builds and iteratively refines a shared slide deck using agents for content planning, visual generation, layout refinement, and grounding verification. In a blind human preference study, \sysname was preferred over both open-source and commercial systems on 77\% of papers while using roughly $4\times$ fewer inference tokens than the strongest open-source baseline. We also introduce ConfArena, an audience-oriented evaluation framework that simulates a conference room and assesses presentations slide by slide. \textsc{ConfArena} matches human system rankings and detects injected presentation problems, including falsified numbers, degraded figures, dropped slides, and shuffled slide order. The code is available at \url{https://github.com/insait-institute/SlideLab}.

\end{abstract}

\section{Introduction}

A conference talk gives an author 10-15 minutes to communicate what a paper develops over eight pages. Unlike a reader, the audience encounters the work for the first time and must follow the presentation as it unfolds. A good slide deck therefore does more than summarize a paper. It selects the right content, presents it in a coherent order, and uses visual elements to help the audience understand the work as the talk progresses.

Recent advances in large language models and agentic frameworks have led to rapid progress in automated slide generation. Existing approaches range from summarization-based methods \citep{d2s,ps5k,doc2ppt} to template-guided editing systems \citep{pptagent,zeng2025slidetailor,slidegen,slidecoder} and end-to-end generation frameworks that use agentic and multi-agent architectures \citep{autopresent,autoslides,deeppresenter} to generate complete presentations from research papers. Despite this progress, generating scientific presentations remains difficult because the system must jointly determine what scientific content should appear on each slide and how it should be presented visually. Recent benchmarks \citep{presentbench,slidesgenbench} show that even state-of-the-art systems still struggle with factual grounding in the source paper, narrative organization, and presentation of visual elements.

The difficulty extends beyond generation to evaluation. Most existing benchmarks \citep{pptagent,presentbench,slidesgenbench,deckbench,arcdeck} ask an LLM to judge the final slide deck against a fixed set of rubrics, such as visual quality, layout consistency, and content fidelity.  While these metrics are useful, they evaluate a completed slide deck rather than the presentation as experienced by an audience. In practice, understanding develops incrementally as a talk progresses, and communication failures naturally surface through audience questions. Existing evaluation frameworks largely overlook this sequential aspect of scientific presentations. 

Motivated by these observations, we address both the generation and evaluation of scientific presentations. We introduce \sysname, an agentic framework that plans a coherent presentation narrative before progressively constructing and refining the slide deck through specialized agents. The framework emphasizes faithfulness to the source paper while improving narrative organization and visual aesthetic. We further introduce a conference-room evaluation environment that evaluates presentations as they are experienced by the audience. A simulated audience follows the presentation slide by slide, asks questions when communication breaks down, and provides a richer assessment than static deck-level evaluation.

Extensive experiments, including human studies with participants spanning undergraduate students to faculty members, show that \sysname outperforms existing open-source systems and closely competes with commercial systems while requiring substantially fewer inference tokens. The proposed evaluation environment also exhibits strong agreement with human judgments, suggesting that it captures aspects of presentation quality overlooked by existing automated metrics.
Our contributions are summarized as follows:
\begin{itemize}
\item We present \sysname, a training-free framework for generating scientific presentations from research papers. The framework progressively constructs and refines the slide deck, producing slide decks with stronger narrative flow, improved visual design, and better factual grounding. In a blind human preference study against three representative open- and closed-source systems, \sysname was preferred for 77\% of the evaluated papers.

\item We introduce \textsc{ConfArena}, a conference-room evaluation environment that assesses scientific presentations through audience interaction rather than static deck evaluation.
\item We will publicly release our codebase for \sysname, ConfArena, web demo, and annotation platform after review.
\end{itemize}

\section{Related Work}
\label{sec:related}
\paragraph{Early Extractive Methods}
Early work formulated slide generation as a document summarization problem~\citep{d2s,ps5k,doc2ppt}, focusing on selecting and compressing salient content from source documents through extraction, ranking, and summarization techniques. These methods established the foundations of automatic slide generation but operate primarily on text, lack multimodal generation capabilities, and predate modern LLM-based systems.

\paragraph{Editing-Based Generation}
A major line of recent work formulates slide generation as adapting or editing existing presentations rather than generating them from scratch~\citep{pptagent,zeng2025slidetailor,slidegen,slidecoder,talk2slides}. These methods leverage reference slide decks, templates, or the deck's underlying object model to inherit content organization and visual design. For example, PPTAgent edits retrieved reference slides, SlideGen extends this paradigm through a multi-agent pipeline for content planning, layout selection, and refinement, and Talk-to-Your-Slides operates directly on a slide's structured object model rather than pixels for cheaper, more precise edits. While these approaches produce visually consistent decks, their dependence on templates or reference presentations constrains layout flexibility, often making generation closer to template filling than true slide synthesis.

\begin{figure*}[t]
    \centering
\includegraphics[width=\textwidth]{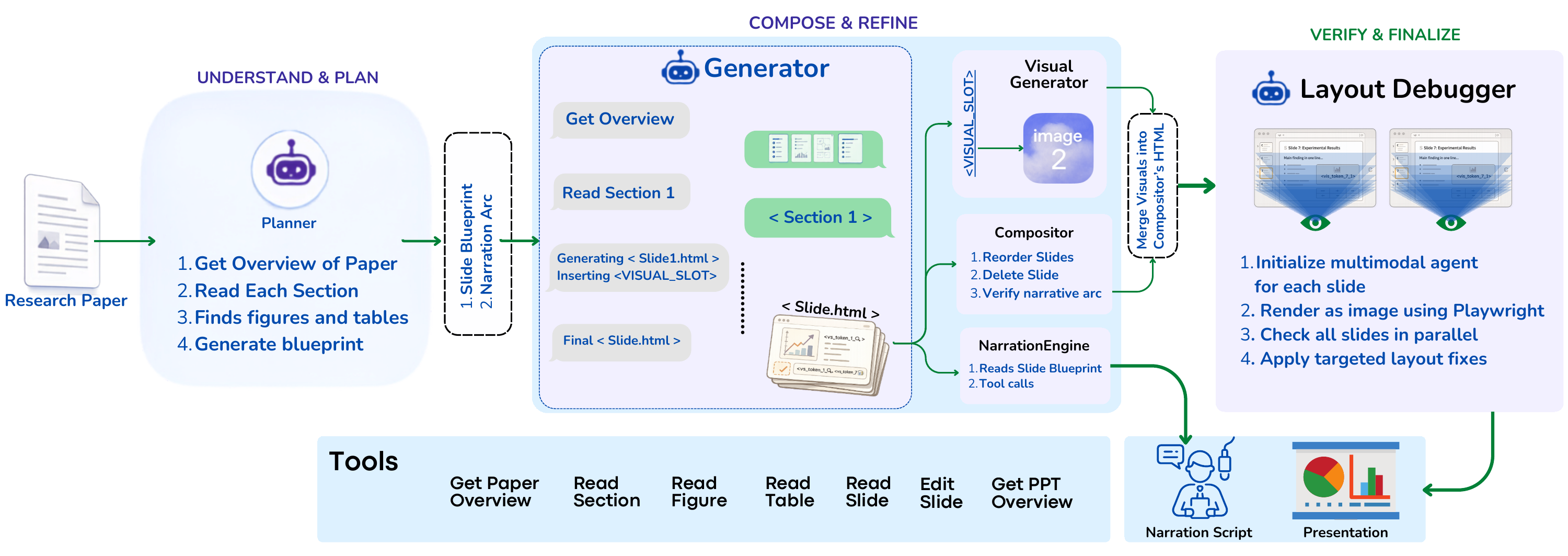}
    \caption{SlideLab Overview: narrative planning, figure design, flow restructuring, layout debugging and grounding.}
    \label{fig:overview}
\end{figure*}

\paragraph{End-to-End Scientific Slide Generation}
Recent work instead aims to synthesize presentations directly from documents or user instructions without relying on reference presentations~\citep{autopresent,autoslides,deeppresenter}. AutoPresent focuses on generating individual slides from natural language instructions but does not explicitly model presentation-level coherence. AutoSlides and DeepPresenter extend this setting to end-to-end scientific presentation generation using multi-stage agentic pipelines, while AeSlides instead targets aesthetic layout quality alone, training with reinforcement learning over verifiable layout metrics rather than content or narrative signals~\citep{aeslides}. Collectively, these methods represent a shift toward fully generative presentation systems that leverage the planning, reasoning, and self-correction capabilities of modern foundation models.

\paragraph{Slide Deck Evaluation}
Existing evaluation methods can be broadly grouped into three categories~\citep{pptagent,presentbench,unippt,arcdeck,slidesgenbench,deckbench,slideaudit}. Judge-based methods (e.g.,\ PPTEval and PresentBench) rely on LLM judges or predefined rubrics to assess completed slide decks, while reference-based benchmarks such as ArcBench and SlidesGen-Bench compare generated presentations against ``ground-truth'' decks. Specialized benchmarks instead target particular failure modes, such as iterative editing (DECKBench), visual design flaws (SlideAudit), or slide-level editing and layout reasoning, where PPTArena scores in-place PowerPoint edits with a dual VLM judge and PPTBench decomposes PowerPoint understanding into Detection, Understanding, Modification, and Generation~\citep{pptarena,pptbench}. Despite these differences, all evaluate presentations as static artifacts with complete deck access. None of them models how audience understanding evolves throughout a presentation or assesses communication in a sequential presentation setting. Our Conference Room environment addresses this gap and is evaluated alongside existing benchmarks rather than replacing them.

\section{\sysname}
\sysname builds upon feedback from human annotations that reveal several flaws in scientific slide generation frameworks, including poor narrative flow, grounding issues, poor layout aesthetics, and high cost (see Figure~\ref{fig:overview}).
Our framework improves the structure, grounding, visual, and other aspects of presentation generation discussed below. It also reduces the cost by 3.6 $\times$ and token consumption by 4$\times$ compared to DeepPresenter.

\subsection{Structured Narrative Planning}

One of the main observations from our human annotation study was that current frameworks still lack factual grounding and fail to create a coherent narrative for the presentation. Although these models can summarize long papers well, they often fail to organize the material into fluent slides. As a result, many generated decks feel out of order, which heavily affects the audience experience.

To address this, we introduce a Planner agent that reads every section, figure, and table of the paper through multi-turn tool calls and generates multiple candidate presentation blueprints. A second agent, powered by GLM-5.2, critiques these candidates and selects the one with the most coherent flow. For each slide, the final blueprint specifies the title, one-sentence takeaway, content source, figures and tables to include, and whether a new visual should be generated. This blueprint is then used by the downstream agents to generate the presentation. Our ablation (Table~\ref{tab:ablate} and \ref{app:model_evaluator_robustness}) shows that removing the Planner causes the largest drop in performance, highlighting the importance of planning the presentation before generating slides.

\subsection{Concept Figure Design}

Scientific presentations often need explanatory diagrams that are not present in the original paper. However, many existing systems either omit such visuals altogether or generate generic images that are only loosely related to the paper. This puts more emphasis on visual appeal than on helping the audience understand the work. As a result, important concepts are often left as blocks of text.

We parse the entire paper and extract all figures, tables, and plots for the downstream agents. The Slide Generator reuses these whenever they are sufficient. If a concept requires a diagram that is not available in the paper, it inserts a \texttt{<!-- VISUAL\_SLOT: description -->} placeholder.

The Visual Generator reads the relevant paper sections together with the slide context and writes a detailed prompt for an image generator (e.g.,\ \texttt{gpt-image-2}), matching the color palette and style already used in the deck. The generated figure is placed into the visual slot, and we check to ensure that it does not overflow the slide or leave large empty gaps, and that it lines up cleanly with any text, tables, or other figures already on the slide. Unlike template-based approaches that force a fixed number of figures or tables onto every slide, we only add a visual when the content actually needs one.

\subsection{Editorial Flow Restructuring}

Generated presentations often contain redundant slides or sections that are better merged or presented in a different order. Existing frameworks handle this only before the generation stage. We instead use a separate post-generation Compositor agent that edits the completed presentation.

Compositor agent can reorder slides, remove redundant ones, merge almost empty slides, move figures to better positions, and insert \texttt{VISUAL\_SLOT}s where a slide becomes too text-heavy. It does not change the design or create new slides. Note that the Compositor has access to the full paper, so every edit is checked against the source before the final deck is returned.

\begin{figure*}[t]
    \centering
    \includegraphics[width=\textwidth]{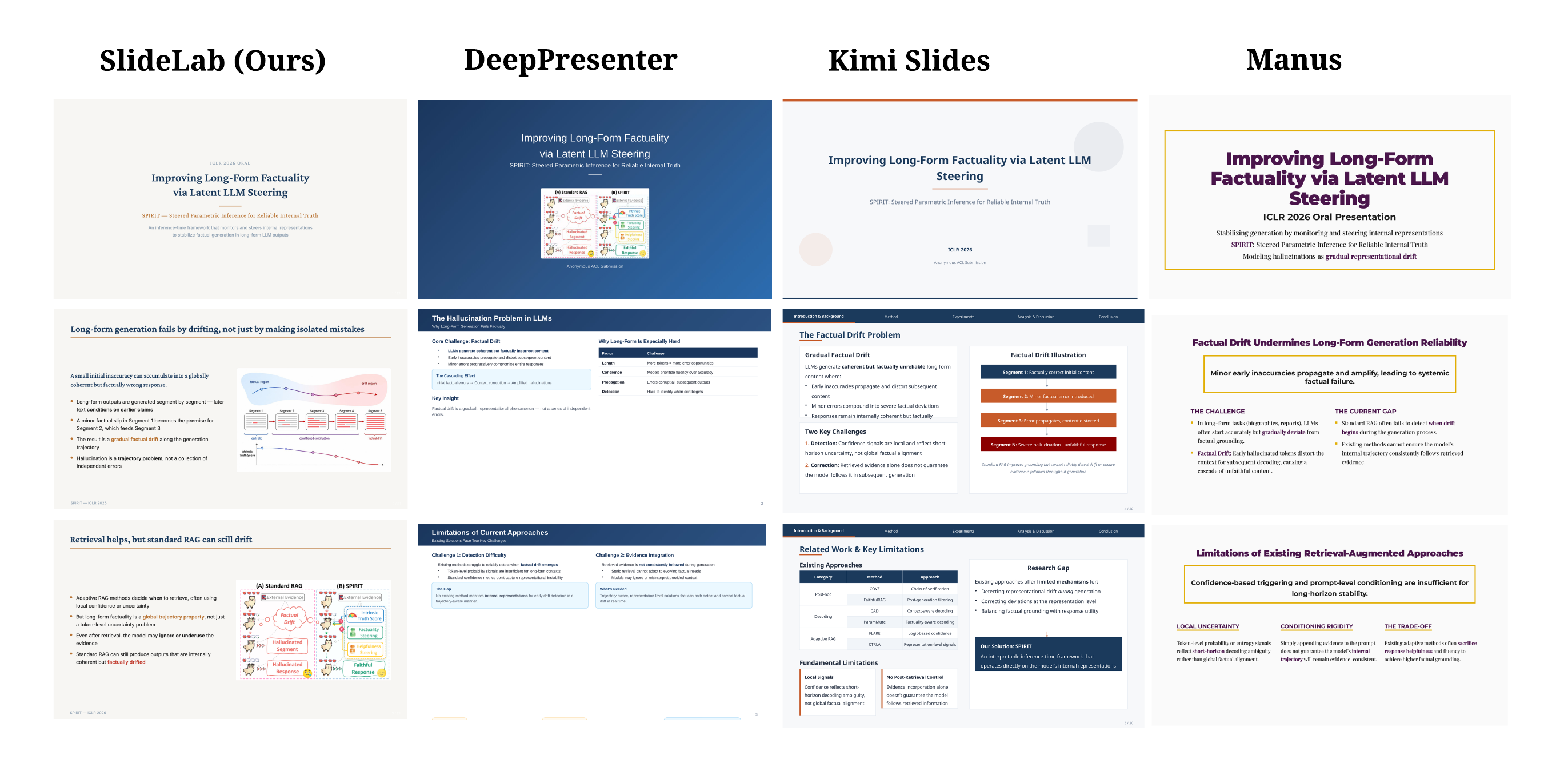}
    \caption{Slides shortcuts comparison between SlideLab (Ours) and DeepPresenter, Kimi Slides, Manus.}
    \label{fig:slide_comp}
\end{figure*}

\subsection{Layout Debugging and Grounding}

After generation, the slides can still have two types of errors: layout defects and grounding errors. Layout problems such as overflow, overlap, clipping, or poor spacing cannot be reliably detected by text only LLMs or simple deterministic methods. To address this, we use a post-generation LayoutDebugger agent powered by a multimodal LLM. Unlike existing frameworks, which rely only on text during generation, our framework performs an explicit visual inspection of every slide. Each slide is rendered as a PNG using Playwright\footnote{Playwright is an open-source browser automation framework developed by Microsoft. \url{https://github.com/microsoft/playwright}.} and passed to the LayoutDebugger. If any layout problem is detected, it returns a corrected HTML version of the slide, which is rendered and checked again for up to three rounds. We then verify grounding using a frontier long-context LLM by providing both the generated HTML and the original paper, allowing it to identify unsupported claims, incorrect numbers, or other factual errors before the presentation is finalized.

\section{ConfArena: Evaluation Environment}

Scientific presentations are intended to communicate research to a diverse conference audience, not merely to showcase visually appealing slide decks. We therefore evaluate presentations in the setting for which they are designed. Instead of relying on a single LLM to score a completed presentation, our environment simulates a conference talk in which a panel of attendees follows the presentation slide by slide, asks questions whenever communication breaks down, and participates in a Q\&A with the presenter. ConfArena then evaluates observable events that occur during a presentation, such as unsupported claims, narrative defects, audience questions, figure failures, and missing scientific content, and aggregates them into complementary evaluation axes, instead of relying on a single holistic judgment.

\begin{table*}[t]
\centering
\small
\setlength{\tabcolsep}{3pt}

\begin{tabular}{lccccccc}
\toprule
& \multicolumn{4}{c}{\emph{Per slide}} & \multicolumn{3}{c}{\emph{Whole talk}} \\
\cmidrule(lr){2-5} \cmidrule(lr){6-8}
System
& \makecell{Grounding.\\errors $\downarrow$}
& Fig.\ errors $\downarrow$
& Design $\uparrow$
& Figure use $\uparrow$
& Narrative.\ errors $\downarrow$
& Coverage $\uparrow$
& Env.\ score $\uparrow$ \\
\midrule

PPTAgent
& 0.98
& 0.64
& 3.21
& 3.18
& 5.40
& 0.61
& 0.42 \\

DeepPresenter
& 2.08
& 0.53
& 3.49
& 3.27
& 5.19
& 0.94
& 0.32 \\

Kimi Slides
& 1.48
& \textbf{0.17}
& 3.67
& 3.08
& 4.56
& 0.91
& 0.58 \\

Manus
& 1.81
& --
& 3.59
& --
& 4.14
& 0.89
& 0.37 \\

\midrule

\sysname\ (ours)
& \textbf{0.71}
& 0.26
& \textbf{4.21}
& \textbf{3.80}
& \textbf{3.29}
& \textbf{0.99}
& \textbf{0.73} \\

\bottomrule
\end{tabular}

\caption{ConfArena results on 100 papers, with three retained decks per paper for SlideLab, DeepPresenter, and PPTAgent, and one for Kimi Slides and Manus. Best values are bold.}

\label{tab:main}
\end{table*}

\subsection{Simulation Overview}
The simulation environment consists of three roles: a \emph{presenter}, an \emph{examiner}, and three \emph{attendee personas}. The presenter has access to the full paper and answers questions as the author, while the attendees see only the slide deck and build their understanding as the presentation progresses. The examiner also reads the paper and serves as a reference for all evaluations. Since the paper, presenter, examiner, and evaluation prompts remain fixed for all slide decks of a paper, differences in the outcome reflect the quality of the slide deck itself.
Each evaluation includes three stages: the presentation, the Q\&A, and the final assessment. Every role is implemented as an independent LLM call. The models and input modalities used at each stage are summarized in \ref{app:confconfig}.

\subsection{Attendees}
A conference audience is inherently diverse. To capture this, we model three attendee personas: an \emph{expert reviewer}, a \emph{learner}, and a \emph{cross-field attendee}. Together, they represent the range of audiences a scientific presentation is intended to serve.

We intentionally avoid elaborate persona prompts. Prior work has shown that increasingly detailed personas can degrade LLM judgment (\citep{kim-etal-2025-persona, zheng-etal-2024-helpful}), while highly specific descriptions risk representing only a narrow stereotype. Each attendee is therefore defined only by three attributes: a \emph{focus}, a \emph{question trigger}, and a short \emph{role description}. The reviewer focuses on scientific correctness and rigor, the learner on whether the methodology and ideas are understandable, and the cross-field attendee on accessibility without specialized background knowledge. Their question triggers determine when something is unclear enough to interrupt the presentation (\ref{app:confconfig}). Consequently, questions arise from distinct audience perspectives rather than a single evaluation criterion, allowing the same presentation to be perceived differently by different attendees, as in a real conference. The attendees evaluate the presentation independently and do not observe each other's reasoning.

\subsection{Presentation}
Before the presentation begins, the examiner reads the paper and prepares a checklist of the key scientific points that should be communicated, such as the motivation, the methodological contribution, and the main experimental findings. These serve as the reference against which the presentation is later evaluated.

The presentation then proceeds slide by slide. For each slide, the examiner compares the rendered slide against the paper to identify unsupported claims, incorrect numerical values, fabricated visuals, and other grounding errors. Independently, each attendee interprets the slide from the perspective of its persona, updates its understanding of the talk, evaluates whether the slide follows naturally from the preceding discussion, and decides whether anything is unclear enough to ask a question. Questions raised during the presentation are collected for the subsequent Q\&A.

\subsection{Q\&A}
The collected questions are posed to the presenter, giving higher priority to questions independently raised by multiple attendees. After the presenter answers using the full paper, the examiner determines why each question arose: because the relevant information was omitted from the slides, communicated unclearly, hidden in an unreadable figure or table, or genuinely beyond the scope of the presentation. The first three cases indicate shortcomings of the presentation, whereas the last reflects discussion that naturally extends beyond the talk itself.

\subsection{Final Assessment}
\label{sec:final_assessment}

Finally, the observations collected throughout the simulation are aggregated into the metrics reported in Table~\ref{tab:main}. \textit{Grounding.\ errors} counts unsupported claims or content, per slide. \textit{Fig.\ errors} counts fabricated or illegible figures per slide. \textit{Design} and \textit{figure use} are scored from 1 to 5, measuring how clean and readable each slide looks and how well its figures support the content. \textit{Narrative.\ errors} counts flow problems across the deck, such as results appearing before the method or a missing conclusion. \textit{Coverage} measures the fraction of the paper's key points covered by the deck (0--1). We rank-normalize each metric within a paper and average them with equal weight to obtain the environment score (0--1).

\section{Experiments}
\label{sec:experiments}

\subsection{Experimental Setup}
\label{sec:exp-setup}
\paragraph{Dataset}

We evaluate on two datasets: (1) the $100$ paper--deck pairs from ArcBench, covering oral-presentation papers from CVPR, ICCV, ICLR, ICML, and NeurIPS (2022--2025), and (2) a held-out set of $30$ machine learning papers. The $100$ paper set is used in automated evaluation runs.

\paragraph{Baselines} We compare against the open-source frameworks PPTAgent and DeepPresenter, along with the commercial systems Kimi Slides~\citep{kimi} and Manus~\citep{manus}. Since PPTAgent edits an existing deck instead of generating one from scratch, it needs a template to start from. We ran it with three different templates for each paper and kept the best score. DeepPresenter is run with default settings, while Kimi Slides and Manus are accessed through their web interfaces using the paper PDF as input. The model configurations for \sysname are detailed in Table~\ref{tab:sysconfig}.

\paragraph{Evaluation}
We evaluate SlideLab through both automatic and human evaluation. For automatic evaluation, we use our proposed ConfArena environment alongside established benchmarks, including PPTEval, PresentBench, and SlidesGen-Bench. Human evaluation is conducted as a blind preference study in which annotators compare complete slide decks without knowing their source. We also assess the quality of ConfArena by measuring its agreement with human judgments and through controlled perturbation experiments that isolate common presentation problems.

\subsection{Human Preference Evaluation}
We conducted a blind human evaluation with 15 volunteer annotators, comprising both authors of the evaluated papers and individuals who were not authors of the annotated papers. The annotators included undergraduates, PhD students, and faculty members. For every paper, annotators were shown four anonymous slide decks (\sysname, DeepPresenter, Kimi Slides, and Manus) in randomized order together with the source paper. Rather than judging visual appearance alone, they were instructed to select the presentation they would use to deliver the paper at a real conference, considering factual correctness, coverage of the paper's main contributions, narrative flow, layout, figure and table usage, and overall presentation quality. 

After selecting the preferred presentation among four, annotators additionally rated the chosen presentation along six dimensions: content grounding, content coverage, narrative structure, visual design, information density, and figure usage, using a five-point Likert scale, to encourage them to consider all aspects of presentation quality rather than making an overall aesthetic judgment. They could optionally provide free-form comments explaining the strengths and weaknesses of the evaluated presentations. The annotation interface and detailed guidelines are provided in the \ref{app}. Table~\ref{tab:human} summarizes the results. Across 30 papers, SlideLab was preferred in 80\% of the annotation (96/120) with 79.16\% pairwise inter-annotator agreement. \sysname was selected for $23$ of the $30$ evaluated papers, while Kimi Slides was preferred for the remaining seven, neither DeepPresenter nor Manus was selected.

The qualitative feedback (Figure~\ref{fig:slide_comp} and Figure~\ref{fig:qualitative}) revealed consistent patterns across the baselines. Manus was often criticized for text-heavy slides and the lack of supporting visual content. In several cases, annotators noted that it inserted screenshots of entire paper pages instead of extracting the relevant figures or tables. DeepPresenter frequently exhibited layout issues, such as excessive whitespace, and its generated figures were often judged unsuitable for conference presentations. It also received the largest number of comments related to factual inaccuracies. Kimi Slides generally produced well-structured presentations but often overcrowded slides with overlapping elements, while receiving the fewest comments about factual errors. In contrast, \sysname consistently avoided these issues, leading annotators to prefer its decks.

\begin{table}[t]
\centering
\small
\begin{tabular}{lcc}
\toprule
System & Papers picked best & \% of 30 \\
\midrule
\sysname (ours) & \textbf{23} & \textbf{77\%} \\
Kimi Slides             & 7  & 23\% \\
DeepPresenter    & 0  & 0\% \\
Manus            & 0  & 0\% \\
\bottomrule
\end{tabular}
\caption{Results of the blind human preference study on 30 research papers. Annotators selected the presentation they would use to deliver the paper at a conference after evaluating factual correctness, coverage, narrative flow, visual design, and figure usage.}
\label{tab:human}
\end{table}

\subsection{Evaluation with ConfArena}
We evaluate on the set of $100$ research papers described in Section~\ref{sec:exp-setup}. PPTAgent, DeepPresenter, and \sysname each generate three independent presentations per paper, giving $900$ decks. Kimi Slides and Manus are web applications with usage limits, so we generate one presentation per paper. Every deck is evaluated independently in ConfArena, and the results are in Table~\ref{tab:main}.

\sysname achieves the highest overall environment score and performs best on all evaluation axes except figure errors. Kimi Slides generated all of the figures programmatically rather than image generation models which helped reduce figure errors. Kimi Slides records the fewest figure errors, but also has noticeably lower coverage of the source paper. DeepPresenter lies at the opposite end, attempting to include more content but accumulating grounding errors together with weaker narrative flow. \sysname achieves the strongest overall performance by maintaining high coverage without sacrificing factual correctness, visual quality, or presentation structure.

\paragraph{\sysname Component Ablations}
We removed each major component of \sysname and re-scored on the $100$-paper set. Table~\ref{tab:ablate} reports the environment score. Removing the Planner causes the largest drop, because the Generator no longer has a structured blueprint. Removing the LayoutDebugger also hurts, mainly through visual-design and illegible-figure penalties. Removing the Compositor or replacing custom visuals with paper figures only causes smaller drops.

\begin{table}[t]
\centering
\small
\begin{tabular}{lc}
\toprule
Configuration & Env.\ score \\
\midrule
\sysname (full)              & \textbf{0.70} \\
$-$ Planner                  & 0.51 \\
$-$ LayoutDebugger           & 0.58 \\
$-$ Compositor               & 0.65 \\
$-$ Custom visuals (paper figs only) & 0.63 \\
\bottomrule
\end{tabular}
\caption{Component ablations of SlideLab. The Planner and LayoutDebugger are the two stages whose removal changes the result significantly. Each configuration is run once per paper on the $100$-paper set, unlike Table~\ref{tab:main}, which averages three independent runs per paper.}
\label{tab:ablate}
\end{table}

\subsection{Performance by Existing Metrics}
We also evaluate the presentations using PPTEval, PresentBench, and SlidesGen-Bench to verify that the improvements observed under ConfArena are reflected by existing evaluation methods.

Table~\ref{tab:extbench} shows a similar trend. \sysname obtains the highest scores on all three PPTEval dimensions and on SlidesGen-Bench, while ranking second on PresentBench, where Kimi Slides performs marginally better. Overall, the rankings produced by existing benchmarks are consistent with those observed under ConfArena, suggesting that the improvements are not biased to the proposed evaluation environment.

\begin{table}[t]
\centering
\footnotesize
\setlength{\tabcolsep}{3pt}
\renewcommand{\arraystretch}{1.1}
\resizebox{\columnwidth}{!}{\begin{tabular}{l ccc c c}
\toprule
 & \multicolumn{3}{c}{PPTEval (1--5)} & PresentBench & SlidesGen-Bench \\
\cmidrule(lr){2-4}
System & Content $\uparrow$ & Design $\uparrow$ & Coherence $\uparrow$ & pass \% $\uparrow$ & quiz acc.\ $\uparrow$ \\
\midrule
PPTAgent      & 3.5 & 3.8 & 3.6 & 63 & 0.74 \\
DeepPresenter & 3.7 & 3.7 & 3.4 & 64 & 0.76 \\
Kimi Slides          & 3.9 & 4.2 & 4.0 & \textbf{73} & 0.79 \\
Manus         & 3.6 & 3.8 & 3.5 & 61 & 0.72 \\
\midrule
\sysname (ours) & \textbf{4.1} & \textbf{4.4} & \textbf{4.2} & 71 & \textbf{0.84} \\
\bottomrule
\end{tabular}}
\caption{Evaluations on the $100$-paper set using other three methods. PPTEval: LLM-judge ratings of content, design, and coherence ($1$--$5$). PresentBench: fraction of decks passing the benchmark's checks (\%). SlidesGenBench: accuracy on quizzes the benchmark generates from the source paper. $\uparrow$ higher is better. Best in bold.}
\label{tab:extbench}
\end{table}

\subsection{Validating ConfArena}
The previous experiments evaluate \sysname using ConfArena. We next validate ConfArena itself by examining whether it aligns with human judgment and whether its individual metrics respond to the presentation failures they are intended to capture.

\paragraph{Does ConfArena Reflect Human Judgment?}
An evaluation framework is only useful if it reflects how people assess presentation quality. We therefore compare the rankings produced by ConfArena and existing automated benchmarks against the blind human study described earlier. Table~\ref{tab:benchmarks} summarizes the resulting system rankings.

Table~\ref{tab:benchmarks} compares the system's rankings and ConfArena's agreement with human deck choices. Humans preferred \sysname, followed by Kimi Slides, while Manus and DeepPresenter tied. ConfArena, PPTEval, and SlidesGen-Bench also rank \sysname first and Kimi Slides second, PresentBench places Kimi Slides first.

\begin{table}[t]
\centering
\footnotesize
\setlength{\tabcolsep}{3pt}
\renewcommand{\arraystretch}{1.1}

\textit{(a) Aggregate system rankings}\par\smallskip
\resizebox{\columnwidth}{!}{\begin{tabular}{lccccc}
\toprule
System & Human & ConfArena & PPTEval & PresentBench & SlidesGenB. \\
\midrule
\sysname      & \textbf{1} & \textbf{1} & \textbf{1} & 2 & \textbf{1} \\
Kimi Slides   & 2 & 2 & 2 & \textbf{1} & 2 \\
Manus         & 3 & 3 & 3 & 4 & 4 \\
DeepPresenter & 3 & 4 & 4 & 3 & 3 \\
\bottomrule
\end{tabular}}

\medskip
\textit{(b) ConfArena agreement with human preferences}\par\smallskip
\begin{tabular}{@{}p{0.78\columnwidth}r@{}}
\toprule
Comparison & Agreement \\
\midrule
Paper-level human majority choice & 78.0\% \\
Individual annotator choice       & 70.5\% \\
\bottomrule
\end{tabular}

\caption{Comparison with human preferences.
(a) Aggregate system rankings ($1$ = best).
Manus and DeepPresenter tie in the human study;
PPTEval ranks use the mean of its three dimensions.
(b) Frequency with which ConfArena selects the same
preferred deck as the human majority for a paper
or as an individual annotator.}
\label{tab:benchmarks}
\end{table}

\begin{table}[t]
\centering
\footnotesize
\setlength{\tabcolsep}{2pt}
\renewcommand{\arraystretch}{1.15}

\begin{tabular}{lcccc}
\toprule
Perturbation & PresentB. & PPTEval & SlidesG. & ConfArena \\
\midrule
Falsified num. & \no & \yes & \no & \yes \\
Degraded fig.  & \yes & \yes & \yes & \yes \\
Dropped slide  & \no & \none & \yes & \yes \\
Shuffled order & \yes & \yes & \yes & \yes \\
\midrule
\emph{Caught /4} & \emph{2} & \emph{3} & \emph{3} & \emph{\textbf{4}} \\
\bottomrule
\end{tabular}
\caption{Can evaluation frameworks detect planted errors in slides? \yes~$=$ the framework's relevant metric moved in the expected direction; \no~$=$ did not; \none~$=$ the framework has no metric targeting that failure.} \label{tab:robustness}
\end{table}

\paragraph{Does ConfArena Measure the Intended Failures?}
A useful evaluation metric should respond to the failure it is intended to measure while remaining relatively insensitive to unrelated modifications. To test this, we manually perturb $15$ high-quality \sysname presentations by introducing targeted errors affecting only a single aspect of the presentation, and then re-evaluate the modified decks using ConfArena.

Table~\ref{tab:perturb} shows that each perturbation primarily affects its corresponding evaluation axis while producing only minor changes in the remaining metrics. For example, shuffling slide order substantially increases narrative defects without affecting grounding, whereas introducing an incorrect numerical value primarily increases grounding errors. Similarly, shrinking figures mainly impacts figure readability, and removing a key methodological slide reduces presentation coverage. These results suggest that the individual ConfArena metrics are sensitive to the presentation failures they are designed to measure rather than unrelated changes.

We also compare this behavior against existing evaluation benchmarks by applying the same perturbations. Table~\ref{tab:robustness} shows the result: ConfArena is the only evaluation that catches all four targeted failures on their intended axes. PPTEval has no metric for missing content at all, while SlidesGen-Bench and PresentBench each have the relevant metric but fail to detect all of the four failures.

\begin{table}[t]
\centering
\small
\setlength{\tabcolsep}{3pt}
\renewcommand{\arraystretch}{1.1}
\resizebox{\columnwidth}{!}{\begin{tabular}{l c c c}
\toprule
 & \multicolumn{3}{c}{Mean change  (positive $=$ worse)} \\
\cmidrule(lr){2-4}
Damage applied & Arc defects & Grounding errors & Figure errors \\
\midrule
Shuffle slide order       & \textbf{+7.25} & +0.11 & -0.01 \\
Inject one false number   & +0.50 & \textbf{+1.51} & +0.20 \\
Shrink figures by 50\%  & +0.25 & -0.17 & \textbf{+0.51} \\
Drop Slides   & \textbf{+2.50} & +0.01 & -0.01 \\
\bottomrule
\end{tabular}}
\caption{We damage $15$ \sysname decks in targeted ways and re-run the ConfArena. Cells show the mean change per metric; bold marks the metric the damage targets. Each damage moves mainly its own metric.}
\label{tab:perturb}
\end{table}

\section{Conclusion}
We presented \sysname, a training-free framework for generating scientific presentations from research papers through structured narrative planning, progressive slide refinement, and multimodal grounding. We also introduced ConfArena, a conference-style evaluation environment that assesses presentations from the audience's perspective through slide-by-slide interactions and Q\&A. Experiments, including blind human evaluation that considers both scientific content and presentation quality, show that SlideLab produces more effective presentations while requiring substantially fewer inference tokens than existing systems. We hope these contributions provide a stronger foundation for future research on both scientific presentation generation and its evaluation.

\section*{Limitations}
Our work focuses on generating presentations from research papers and evaluates systems within this setting. While the \sysname is largely domain-agnostic, the experiments in this work are limited to conference style presentations. Many of the design principles used by \sysname, including narrative planning, grounding, and iterative slide refinement, are not specific to scientific presentations and may generalize to other presentation domains. Future work could investigate the applicability of the framework to educational, business, and technical presentations, which often differ in audience, communication objectives, and presentation style.

\section*{Ethical Considerations}
Our work is intended to support scientific communication by assisting in the preparation and evaluation of research presentations, not to replace human authorship or scientific judgment. Since the system can generate incorrect or misleading content, all generated slides should be reviewed by authors before public use.

\bibliography{custom}

\appendix

\input{appendix}

\end{document}

%% file: appendix.tex
\section{Appendix}
\label{sec:appendix}

\subsection{\sysname Configuration}
\label{app:sysconfig}
Table~\ref{tab:sysconfig} lists the model used at each stage of \sysname, with max number of multi-turn rounds. All text agents run at temperature $1.0$ with high reasoning effort. The Planner, Slide Generator, and Compositor each run as a tool-use loop with the round limits shown, hitting the limit forces the agent to produce its final output from what it has gathered so far. 

\begin{table*}[!t]
\centering
\small
\begin{tabularx}{\textwidth}{
    @{}
    p{0.20\textwidth}
    X
    p{0.20\textwidth}
    p{0.18\textwidth}
    @{}
}
\toprule
\textbf{Stage}
& \textbf{Sub-stage}
& \textbf{Model}
& \textbf{Max rounds / calls} \\
\midrule

\textbf{Planner}
& \textit{Candidate plan generation}
& GPT-5.5
& 60 \\

& \textit{Candidate critic and selector}
& GLM-5.2
& 1 \\

& \textit{Plan grounding verifier}
& GPT-5.5
& 1 \\

\addlinespace
\textbf{Slide Generator}
& \textit{Slide generation}
& MiMo-v2.5-Pro
& 50 \\

\addlinespace
\textbf{Visual Generator}
& \textit{Visual instruction / prompt generation}
& MiMo-v2.5-Pro
& 15 \\

& \textit{Image generation}
& GPT-Image-2
& 1 per generated visual \\

\addlinespace
\textbf{Compositor}
& --
& MiMo-v2.5-Pro
& 30 \\

\addlinespace
\textbf{Layout Debugger}
& --
& GPT-5.5
& 3 per slide \\

\addlinespace
\textbf{Narration Engine}
& --
& MiMo-v2.5-Pro
& 15 \\

\bottomrule
\end{tabularx}

\caption{Models and maximum round 
each \sysname stage.}
\label{tab:sysconfig}
\end{table*}

\subsection{Model Configuration and Ablation Results}
\label{app:model_evaluator_robustness}

For PPTAgent and DeepPresenter, we use their default configurations with GPT-5.5 as the LLM. Additionally, DeepPresenter uses GPT-Image-2 for image generation. In the experiments reported above, MiMo-V2.5-Pro serves as the primary model for slide generation, visual prompting, and deck composition, while GPT-5.5 is only used in a supporting role for planning and layout debugging.

To further examine the contribution of the architecture under a shared model configuration, we run the component ablations on 30 papers using MiMo-V2.5-Pro throughout the pipeline, including planning, layout debugging, and grounding verification. We evaluate each ablation with ConfArena using two independent evaluator models, GPT-5.4 and Gemini-3.8-Flash to mitigate evaluator-specific bias.

We then remove each SlideLab component separately
while keeping the MiMo-V2.5-Pro configuration.
Table~\ref{tab:model_ablation} shows that every
removal lowers the score under both judges.
Removing the Planner causes the largest decrease:
from 0.74 to 0.53 under GPT-5.4 and from 0.78 to
0.62 under Gemini. This supports the importance
of planning the presentation before generating
slides, even when planning and generation use
the same model.

\begin{table}[!htbp]
\centering
\small
\begin{tabularx}{\columnwidth}{@{}Xcc@{}}
\toprule
& \multicolumn{2}{c}{\textbf{ConfArena score} $\uparrow$} \\
\cmidrule(lr){2-3}
\textbf{Configuration}
& \textbf{GPT-5.4}
& \makecell{\textbf{Gemini-3.8-}\\\textbf{Flash}} \\
\midrule

\makecell[l]{\textbf{\sysname (full)}\\
\textit{(MiMo-V2.5-Pro)}}
& \textbf{0.74}
& \textbf{0.78} \\

\addlinespace[4pt]
\textcolor{red!70!black}{$-$} Planner
& 0.53 & 0.62 \\

\textcolor{red!70!black}{$-$} LayoutDebugger
& 0.60 & 0.69 \\

\textcolor{red!70!black}{$-$} Compositor
& 0.70 & 0.74 \\

\textcolor{red!70!black}{$-$} Custom visuals
(paper figures only)
& 0.69 & 0.70 \\

\bottomrule
\end{tabularx}

\caption{Component ablations on 30 papers using
MiMo-V2.5-Pro for all text agents. Columns report
environment scores under two ConfArena runs with different model.
Each minus denotes removal of one component
from the full pipeline.}
\label{tab:model_ablation}
\end{table}

\begin{table*}[!t]
\centering
\small
\begin{tabularx}{\textwidth}{@{}Xccccccc@{}}
\toprule
& \multicolumn{4}{c}{\textit{Per slide}}
& \multicolumn{3}{c}{\textit{Whole talk}} \\
\cmidrule(lr){2-5}
\cmidrule(lr){6-8}

\textbf{System}
& \makecell{\textbf{Grounding}\\\textbf{errors} $\downarrow$}
& \makecell{\textbf{Figure}\\\textbf{errors} $\downarrow$}
& \textbf{Design} $\uparrow$
& \makecell{\textbf{Figure}\\\textbf{use} $\uparrow$}
& \makecell{\textbf{Narrative}\\\textbf{errors} $\downarrow$}
& \textbf{Coverage} $\uparrow$
& \makecell{\textbf{Env.}\\\textbf{score} $\uparrow$} \\
\midrule

\makecell[l]{PPTAgent 
\textit{(GPT-5.5)}}
& 1.462 & 0.443 & 3.441 & 3.090
& 5.13 & 0.64 & 0.38 \\

\addlinespace
\makecell[l]{DeepPresenter \textit{(GPT-5.5)}}
& 1.882 & 0.316 & 3.725 & 3.368
& 4.77 & 0.753 & 0.30 \\

Kimi Slides
& \textbf{1.162} & 0.156 & 4.430 & 2.830
& 3.94 & 0.924 & 0.61 \\

Manus
& 1.650 & -- & 3.990 & --
& 3.20 & 0.691 & 0.31 \\

\midrule
\textbf{\sysname (ours)}\newline
\textit{(MiMo-V2.5-Pro, all text stages)}
& 1.244
& \textbf{0.140}
& \textbf{4.673}
& \textbf{4.525}
& \textbf{2.400}
& \textbf{0.927}
& \textbf{0.784} \\

\bottomrule
\end{tabularx}

\caption{System comparison on 30 human-annotated papers using
Gemini-3.8-Flash for all ConfArena evaluation roles.
SlideLab uses MiMo-V2.5-Pro for all text agents.
Metric definitions follow Table~\ref{tab:main}.
Bold marks the best value in each column; dashes
indicate unavailable figure metrics.}
\label{tab:gemini_comparison}
\end{table*}

\subsection{ConfArena Configuration}
\label{app:confconfig}
\begin{figure*}[!t]
\centering
\includegraphics[width=\textwidth]{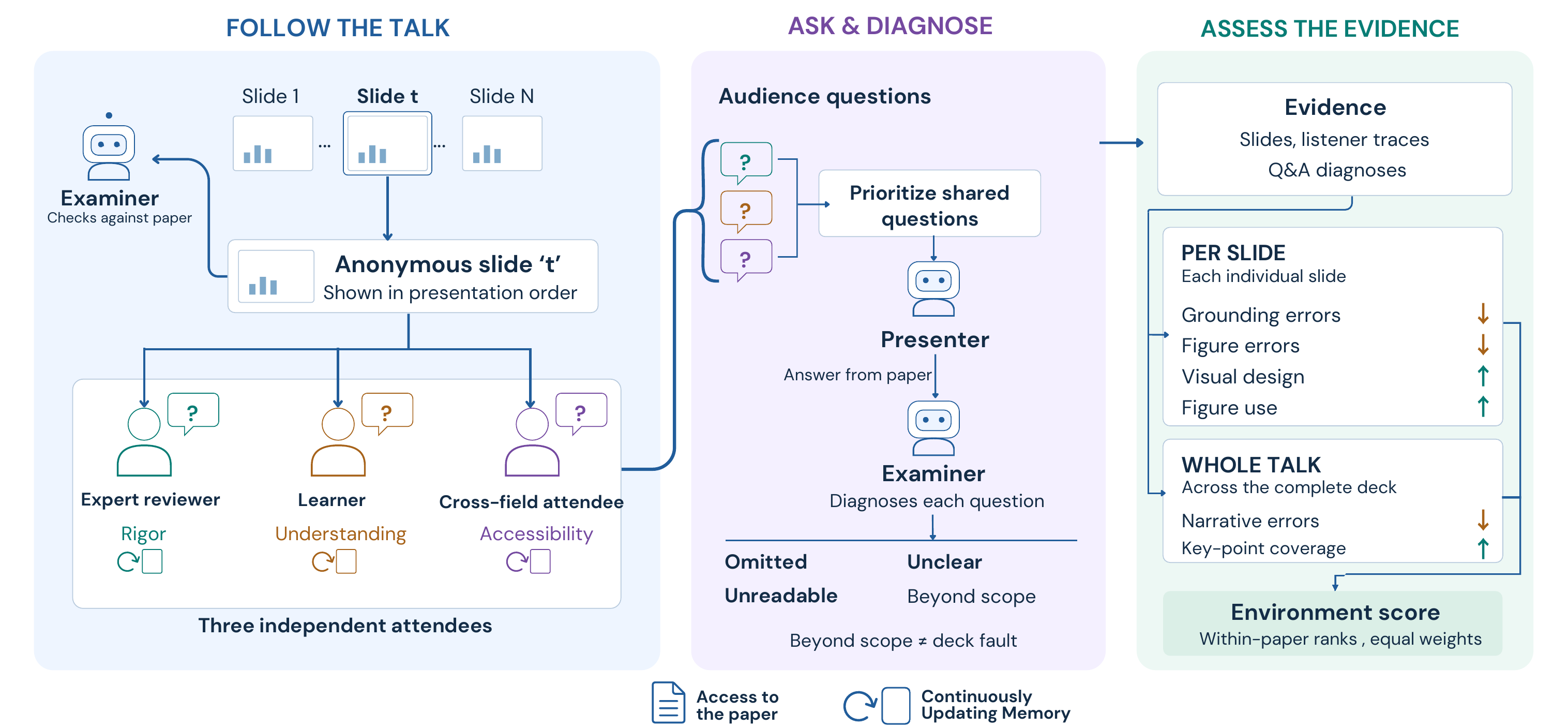}
\caption{ConfArena evaluation architecture: sequential slide review by the examiner and three independent attendees, audience Q\&A and question diagnosis, and aggregation of per-slide and whole-talk evidence into the environment score.}
\label{fig:confarena_architecture}
\end{figure*}

Figure~\ref{fig:confarena_architecture} summarizes the sequential presentation, audience Q\&A, and evidence-based assessment in ConfArena.

ConfArena runs each deck through a simulated conference talk with three independent attendee personas, a reviewer, a learner, and a cross-field listener, each tracking its own understanding of the presentation as it proceeds. The main-paper experiments use GPT-5.4 for all ConfArena evaluation calls. To examine possible evaluator-model bias, we additionally report
results using Gemini-3.8-Flash with medium
reasoning effort. 
Table~\ref{tab:gemini_comparison} reports the full system comparison under Gemini-3.8-Flash while Table~\ref{tab:model_ablation} compares component ablations under both evaluators. SlideLab achieves
the highest environment score despite using MiMo-V2.5-Pro throughout its text-agent pipeline, while PPTAgent and DeepPresenter use GPT-5.5. SlideLab leads on five of the six individual metrics with
Kimi Slides having the fewest grounding errors.
Only the per-slide visual pass sees the rendered slide image, every other call works from slide text and the source paper.

\begin{table}[!htbp]
\centering
\small
\begin{tabularx}{\columnwidth}{@{}p{0.24\columnwidth}X@{}}\toprule
\textbf{Attendee} & \textbf{Focus and Question Trigger} \\
\midrule
Expert reviewer &
Focuses on scientific rigor and correctness. Asks a question when a claim, result, or numerical value appears unsupported, inconsistent, or overstated. \\

Learner &
Focuses on whether the core ideas, methodology, and motivation are easy to follow. Asks a question when an important concept or step is insufficiently explained. \\

Cross-field attendee &
Focuses on accessibility without deep subfield knowledge. Asks a question when jargon or field-specific assumptions are introduced without adequate explanation. \\
\bottomrule
\end{tabularx}
\caption{Attendee personas used in ConfArena.}
\label{tab:personas}
\end{table}

\subsection{Evaluation Configs}
\label{app:protocol}
PPTEval, PresentBench, and SlidesGen-Bench are run with the default settings from their released codebases with openai/gpt-5.4 as the judge model. Decks average 19.4 slides for SlideLab, 18.5 for DeepPresenter, 19.7 for Kimi Slides, and 14.2 for Manus.

\subsection{Sequential Evaluation and Audience Ablations}
\label{app:sequential_eval}

\paragraph{Comparison with Separate Slide Scoring} We compare ConfArena with a static examiner that scores each slide separately, without following the deck in sequence. Both ConfArena and the static examiner use Gemini-3.8-Flash on the same 30 papers. 

Table~\ref{tab:static_corr} shows that their scores are closely correlated for grounding, figure errors, design, and figure use but the correlations are lower for coverage and narrative errors, showing that ConfArena assesses deck-level content and flow differently. The narrative results explain this difference (Table~\ref{tab:static_narrative}). The static examiner reports slightly fewer narrative errors for Kimi Slides than for SlideLab (0.80 versus 0.90). ConfArena reports fewer errors for SlideLab (2.40 versus 3.94).

\begin{table}[!htbp]
\centering
\small
\begin{tabularx}{\columnwidth}{@{}Xc@{}}
\toprule
\textbf{Metric} & \textbf{Pearson correlation ($r$)} \\
\midrule
Grounding errors & 0.939 \\
Figure errors    & 0.952 \\
Design           & 0.981 \\
Figure use       & 0.917 \\
Coverage         & 0.637 \\
Narrative errors & 0.552 \\
\bottomrule
\end{tabularx}
\caption{Correlation between ConfArena scores and
scores from an examiner that evaluates slides
separately. Both use Gemini-3.8-Flash.}
\label{tab:static_corr}
\end{table}

\begin{table}[!htbp]
\centering
\small
\begin{tabularx}{\columnwidth}{@{}Xcc@{}}
\toprule
& \multicolumn{2}{c}{\textbf{Narrative errors} $\downarrow$} \\
\cmidrule(lr){2-3}
\textbf{Evaluator} & \textbf{\sysname} & \textbf{Kimi Slides} \\
\midrule
Static examiner & 0.90 & 0.80 \\
ConfArena       & 2.40 & 3.94 \\
\bottomrule
\end{tabularx}
\caption{Mean narrative errors per deck under the
static examiner and ConfArena.}
\label{tab:static_narrative}
\end{table}

\paragraph{Audience Roles and Q\&A}
We examine how attendee roles and Q\&A affect ConfArena's narrative assessment. We compare the full evaluator with three alternatives: neutral attendee roles, no Q\&A, and an examiner-only configuration without attendees. Table~\ref{tab:evaluator_ablation} shows that each change reduces the number of narrative problems reported. Removing Q\&A or using only the examiner has a larger effect than replacing the attendee roles.

\begin{table}[!htbp]
\centering
\small
\begin{tabularx}{\columnwidth}{@{}cXr@{}}
\toprule
\multicolumn{2}{@{}l}{\textbf{Evaluator condition}}
& \makecell[r]{\textbf{Narrative errors}\\
\textbf{per deck}} \\
\midrule

\faUsers
& \textbf{Full ConfArena}
& \textbf{3.88} \\

\addlinespace
\faMeh
& Neutral attendee roles
& 3.40 \\

\addlinespace
\textcolor{red!70!black}{\faCommentSlash}
& Without Q\&A
& 2.30 \\

\addlinespace
\faUser
& Examiner only
& 1.75 \\

\bottomrule
\end{tabularx}

\caption{Mean narrative errors per deck under
different ConfArena configurations.}
\label{tab:evaluator_ablation}
\end{table}
\subsection{Computational Cost}
Table~\ref{tab:cost} breaks down the average wall-clock time and API cost per deck on the $100$-paper set. The full system costs approximately \$$0.58$ per paper and runs in under twelve minutes. \sysname uses about ~$700$K tokens per deck, roughly $4\times$ fewer than DeepPresenter ($\sim$$3000$K), mainly because the Planner reads the paper once and the Slide Generator works slide by slide against a fixed blueprint, instead of reflecting over the whole deck repeatedly.

\begin{table}[!htbp]
\centering
\small
\begin{tabularx}{\columnwidth}{@{}Xccc@{}}
\toprule
\textbf{Stage} & \makecell{\textbf{Time}\\\textbf{(s)}} & \makecell{\textbf{Cost}\\\textbf{(\$)}} & \makecell{\textbf{Tokens}\\\textbf{(M)}} \\
\midrule
Planner                          & 96  & 0.10 & 0.09 \\
Slide Generator                  & 264 & 0.03 & 0.23 \\
ImageVisual\allowbreak Generator + Compositor & 261 & 0.42 & 0.32 \\
LayoutDebugger                   & 122 & 0.03 & 0.05 \\
\midrule
\sysname total                   & \textbf{743} & 0.58 & 0.69 \\
DeepPresenter                    & 1620 & 2.1 & 2.9 \\
\bottomrule
\end{tabularx}
\caption{Average wall-clock time (end to end), API cost, and token usage per deck.}
\label{tab:cost}
\end{table}

\subsection{Human Evaluation Interface}
\label{app}
\begin{figure*}[!t]
\centering
\includegraphics[width=0.9\textwidth]{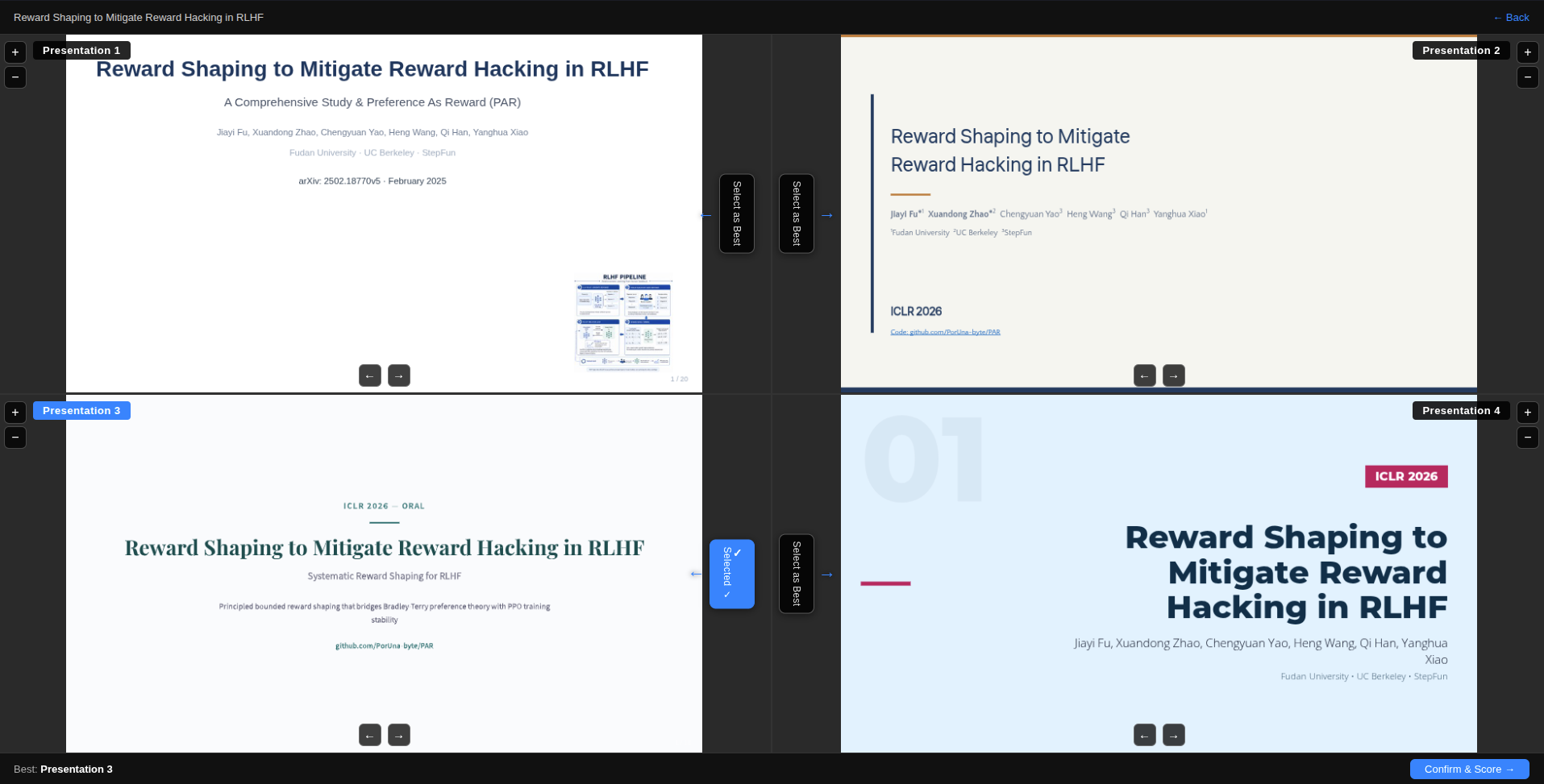}
\caption{Annotation interface for the human preference study. Annotators see the four anonymized decks side by side, pick the one they would present, then rate it on six dimensions.}
\label{fig:annotation_ui}
\end{figure*}

Figure~\ref{fig:annotation_ui} shows the annotation interface used in the human preference study. Annotators first pick the deck they would use to present the paper at a conference, then rate that deck on six dimensions using a five-point scale. The six dimensions and their anchors are listed in Table~\ref{tab:guidelines}. Annotators were known to the authors and participated on a voluntary basis. They were aware that their annotations will be shown for analysis in our work.

\begin{table}[!htbp]
\centering
\small
\begin{tabularx}{\columnwidth}{@{}p{0.34\columnwidth}X@{}}
\toprule
\textbf{Dimension} & \textbf{What annotators are asked} \\
\midrule
Content Grounding & Do the slides match what the paper says? \\
Content Coverage & Does the deck cover the paper's key contributions? \\
Narrative Structure & Do the slides tell a coherent story? \\
Visual Design & Does it look like a professional conference talk? \\
Information Density & Are slides concise, not walls of text? \\
Figure Usage & Are figures well-chosen and sized correctly? \\
\bottomrule
\end{tabularx}
\caption{The six dimensions annotators rate after picking their preferred deck. Each uses a five-point scale with the anchors shown in the interface.}
\label{tab:guidelines}
\end{table}

\paragraph{Why Likert Scoring?}
Many presentation attributes, including visual design, figure use, and narrative quality, cannot be adequately captured by binary decisions. We therefore use Likert ratings for these dimensions, allowing the evaluation to capture incremental differences in presentation quality that binary labels would overlook.

\subsection{Detailed Perturbation Scores}
\label{app:perturb_detail}
\begin{table*}[!t]
\centering
\small
\begin{tabular}{ll l rr r c}
\toprule
\textbf{Perturbation} & \textbf{Benchmark} & \textbf{Metric (direction)} & \textbf{Baseline} & \textbf{Perturbed} & $\Delta$ & \textbf{Detected?} \\
\midrule
Falsified number
  & PPTEval        & content (1--5, $\uparrow$)            & 4.30 & 4.27 & $-0.03$  & \yes \\
  & SlidesGen-Bench & quiz accuracy (0--1, $\uparrow$)      & 0.98 & 0.98 & $+0.00$  & \no \\
  & PresentBench    & material-dep.\ pass \% ($\uparrow$)   & 27.5 & 45.0 & $+17.5$  & \no \\
  & ConfArena        & faithfulness (0--1, $\uparrow$)       & 0.58 & 0.57 & $-0.01$  & \yes \\
\midrule
Degraded figure
  & PPTEval        & design (1--5, $\uparrow$)             & 4.05 & 3.85 & $-0.20$  & \yes \\
  & SlidesGen-Bench & visual weighted total (0--10, $\uparrow$) & 8.45 & 8.22 & $-0.23$ & \yes \\
  & PresentBench    & material-indep.\ pass \% ($\uparrow$) & 100.0 & 92.0 & $-8.0$   & \yes \\
  & ConfArena        & figure legible \% ($\uparrow$)        & 75.3 & 74.2 & $-1.06$  & \yes \\
\midrule
Dropped slide
  & PPTEval        & coverage axis                          & --   & --   & --       & \none \\
  & SlidesGen-Bench & quiz accuracy (0--1, $\uparrow$)      & 0.98 & 0.97 & $-0.01$  & \yes \\
  & PresentBench    & material-dep.\ pass \% ($\uparrow$)   & 27.5 & 27.5 & $+0.00$  & \no \\
  & ConfArena        & coverage (0--1, $\uparrow$)           & 0.92 & 0.90 & $-0.02$ & \yes \\
\midrule
Shuffled order
  & PPTEval        & coherence (1--5, $\uparrow$)          & 4.90 & 3.70 & $-1.20$  & \yes \\
  & SlidesGen-Bench & logical flow (1--10, $\uparrow$)      & 8.80 & 6.40 & $-2.40$  & \yes \\
  & PresentBench    & material-indep.\ pass \% ($\uparrow$) & 100.0 & 90.0 & $-10.0$  & \yes \\
  & ConfArena        & narrative (0--1, $\uparrow$)          & 0.93 & 0.69 & $-0.238$ & \yes \\
\bottomrule
\end{tabular}
\caption{Mean baseline and perturbed scores per benchmark per perturbation, across $15$ papers. $\Delta$ = perturbed minus baseline; for metrics where higher is better, a negative $\Delta$ means the benchmark moved in the expected direction. \yes~$=$ detected, \no~$=$ not detected, \none~$=$ no metric exists for that failure mode. Baselines are the mean over the $15$ unperturbed decks.}
\label{tab:perturb_detail}
\end{table*}

Table~\ref{tab:perturb_detail} reports the full numeric results behind Table~\ref{tab:robustness}. For each benchmark and perturbation we show the metric that benchmark uses for that failure mode, its mean value on the unperturbed decks, its mean value on the perturbed decks, and the change. A metric counts as detecting the failure if it moves in the expected direction (down for scores where higher is better, up for error counts). Cells marked -- mean the benchmark has no metric for that failure mode.

For ConfArena, we compute faithfulness and narrative scores from grounding and narrative error counts as follows:
\[
\begin{aligned}
\text{Faithfulness} &= \exp(-g),\\
\text{Narrative} &= \exp\left(-\frac{n}{s}\right),
\end{aligned}
\]

where $g$ is the mean grounding error count per slide, $n$ is the total narrative error count, and $s$ is the number of slides. Figure legibility is the percentage of figures judged readable, covering only the readability component of figure errors. Coverage has the same definition as in Table~\ref{tab:main}.

\paragraph{Figure Errors and Figure Use}
\sysname uses native image-generation models (GPT-Image-2 in current setup), which can make mistakes when generating complex visuals. Kimi Slides instead generates diagrams and charts programmatically, avoiding image models and this may help limit figure errors. Because of this, Kimi Slides has fewer figure errors in ConfArena (0.17 vs. 0.26), Table~\ref{tab:main}. But its visuals, however, are typically limited to geometric plots, charts, and simple block diagrams. During annotation, annotators often observed that these visuals were clean and error-free, but hardly represented the paper. SlideLab can generate new paper specific visuals/images when needed. This relevance is also reflected by the figure-use axis where figures generated by Kimi Slides are scored 3.08 Vs the 3.80 of \sysname. Kimi Slides reduces figure errors through code based simple visuals, while SlideLab can produce visuals that more directly represent and explain the paper.
In the additional (Table~\ref{tab:gemini_comparison}), SlideLab has slightly fewer figure errors than Kimi Slides (0.140 versus 0.156), while retaining a higher figure-use score. Here, the figure-use advantage remains.

\clearpage
\begin{figure*}[p]
    \centering
    \includegraphics[width=0.97\textwidth]{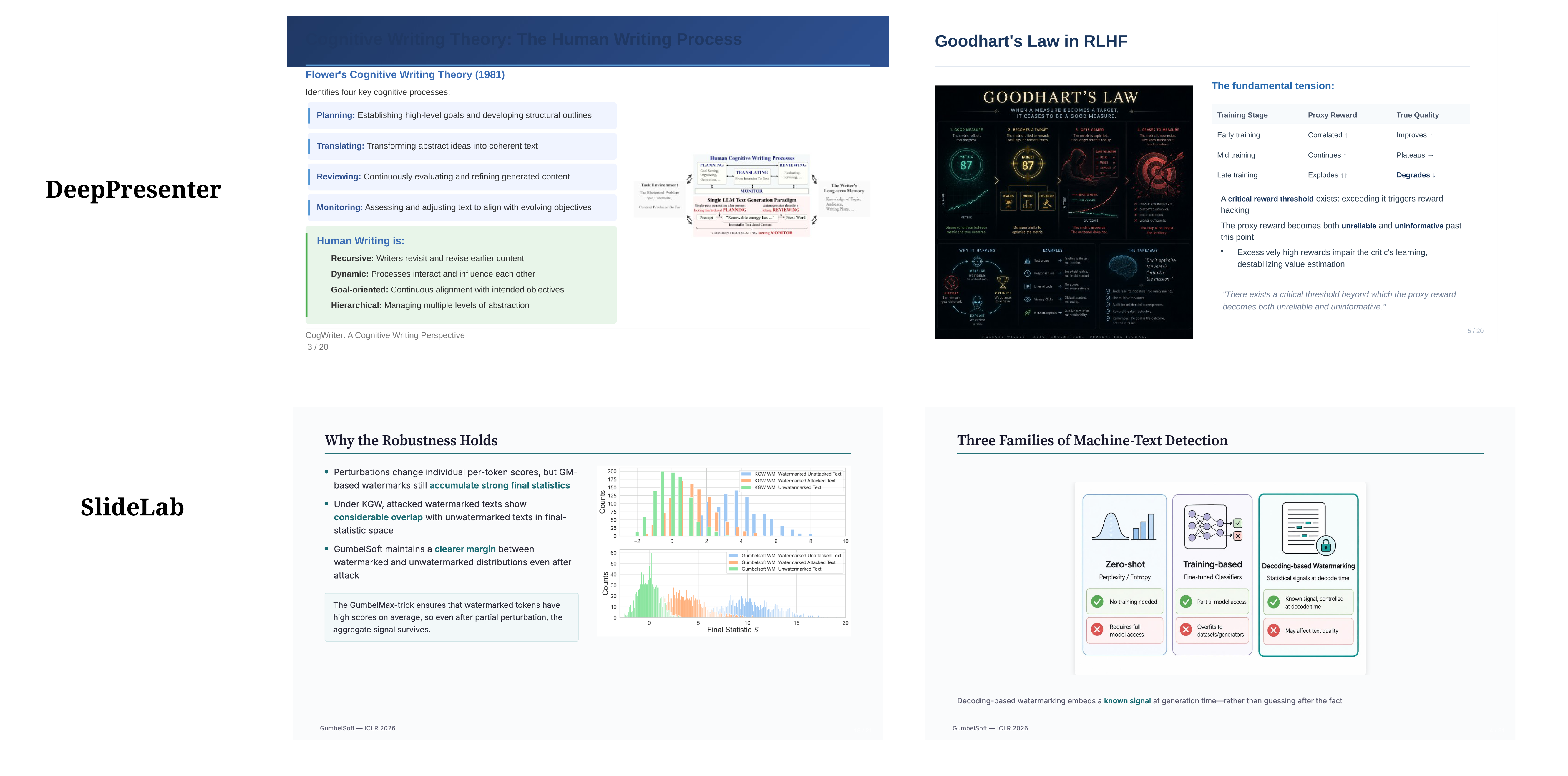} \\
    \includegraphics[width=0.97\textwidth]{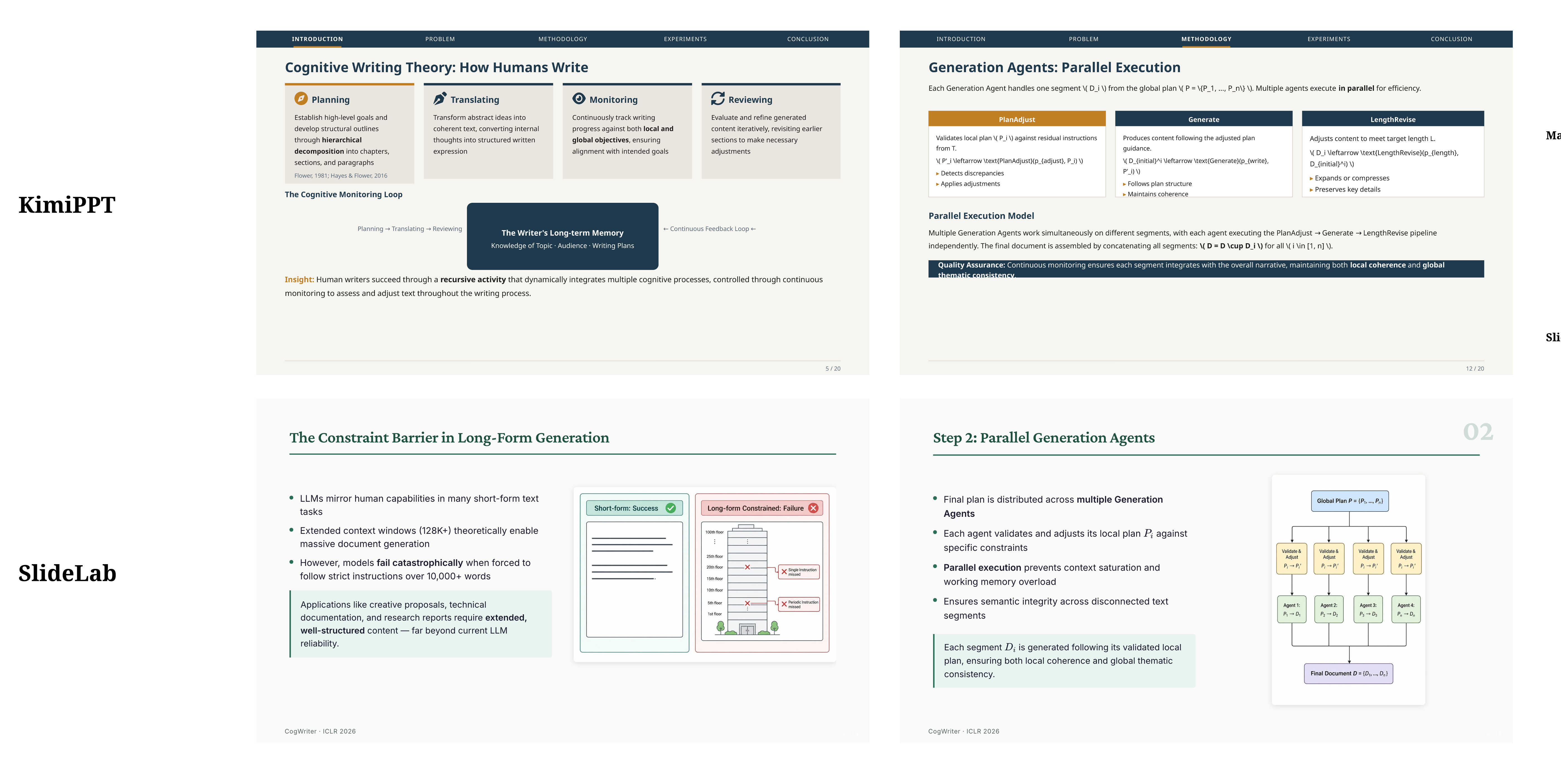} \\
    \includegraphics[width=0.97\textwidth]{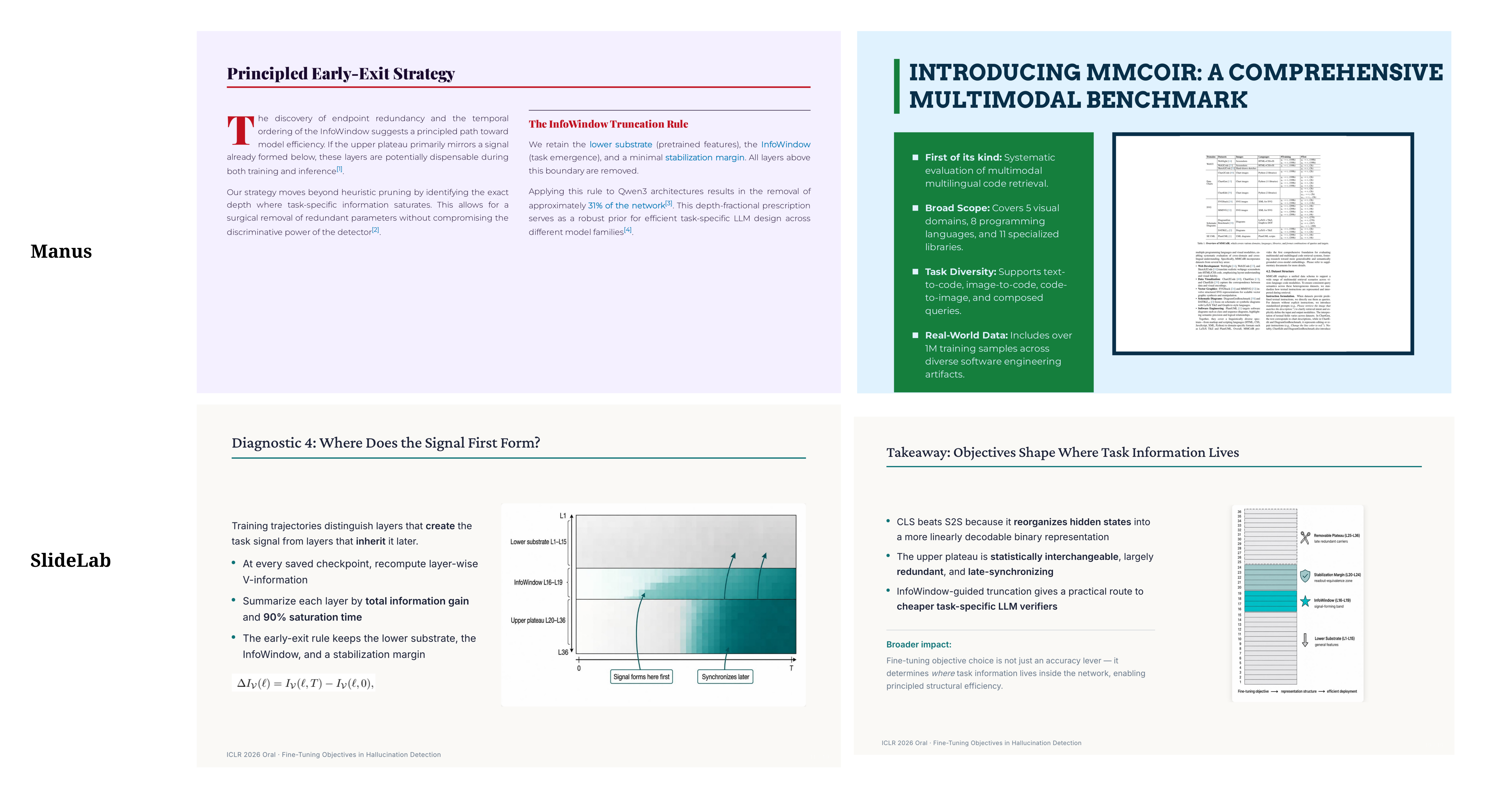}

    \caption{Representative qualitative comparisons supporting the quantitative results reported in the main paper.}
    \label{fig:qualitative}
\end{figure*}